\documentclass[10pt]{article} 

\usepackage[preprint]{rlj} 

\usepackage{amssymb}            
\usepackage{mathtools}          
\usepackage{mathrsfs}           
\usepackage{graphicx}           
\usepackage{subcaption}         
\usepackage[space]{grffile}     
\usepackage{url}                
\usepackage{lipsum}  
\usepackage{cancel}

\title{Physical Reinforcement Learning}

\setrunningtitle{Physical Reinforcement Learning}

\author{Sam Dillavou\textsuperscript{1}, Shruti Mishra\textsuperscript{2}\textsuperscript{*}}

\emails{dillavou@upenn.edu, \ \ sm3056@cam.ac.uk}

\affiliations{
$^{1}$\textbf{Department of Physics and Astronomy, University of Pennsylvania, USA}\\
$^{2}$\textbf{Department of Applied Mathematics and Theoretical Physics, University of Cambridge, UK\thanks{Work done while at Sony AI, Europe.}}\\
}

\begin{document}

\maketitle  
\renewcommand{\thefootnote}{*}
\footnotetext{work done while at Sony AI}

\begin{abstract}
Digital computers are power-hungry and largely intolerant of damaged components, making them potentially difficult tools for energy-limited autonomous agents in uncertain environments. Recently developed \textit{Contrastive Local Learning Networks} (CLLNs) — analog networks of self-adjusting nonlinear resistors — are inherently low-power and robust to physical damage, but were constructed to perform supervised learning. In this work we demonstrate success on two simple RL problems using Q-learning adapted for simulated CLLNs. Doing so makes explicit the components (beyond the network being trained) required to enact various tools in the RL toolbox, some of which (policy function and value function) are more natural in this system than others (replay buffer). We discuss assumptions such as the physical safety that  digital hardware requires, CLLNs can forgo, and biological systems cannot rely on, and highlight secondary goals that are important in biology and trainable in CLLNs, but make little sense in digital computers.
\end{abstract}

\section{Introduction}
\label{sec:intro}
\renewcommand{\thefootnote}{1}
Digital computers are powerful and versatile machines, but have significant weaknesses, especially in areas relevant to reinforcement learning (RL). First, they are fault intolerant and thus susceptible to damage. Disable a handful or even a single transistor in a CPU or GPU and it may crash the entire system\footnote{https://www.ntchip.com/electronics-news/transistors-in-cpu}. Having a {single} transistor erroneously flip even on 1\% of write operations could catastrophically disable the entire machine. This is because the function of each component is intrinsically tied to its location within the system, and the entire operation relies on (nearly) error-free operation. Second, they are power-hungry, with laptop CPUs consuming approximately 50~W. This stems from the high energy cost of maintaining `perfect' operation, as well as the shuttling of data between processing and memory.

Low-power operability and fault tolerance are areas where biological systems thrive, as they are autonomous agents in an often energy-scarce and dangerous world. Brains can take significant damage and continue to function, including destruction of single-neurons \citep{wang_laser_2014}, traumatic brain injuries \citep{chua_brief_2007}, and even removal of large brain regions \citep{granovetter_childhood_2022}. This robustness stems from brains' distributed processing and emergent, rather than linear, computation. The human brain is also energy efficient, using 20~W in total \citep{balasubramanian_brain_2021} while performing perception, cognition, motor control, homeostatic functions, learning, and more. This energy efficiency is in part due to the overlap of memory and computation, and to \textit{natural primitives} \citep{mead_neuromorphic_1990}, whereby low-level computations utilize low-power analog physical or chemical processes rather than digital logic.

In spite of the potential advantages, there are few instances of artificial and non-digital hardware performing RL tasks \citep{mak_currentmode_2007,mak_cmos_2010}. Many instances of digitally-enhanced or simulated analog systems have been used in an RL setting (as in \citet{mikaitis_neuromodulated_2018}), but few if any hardware demonstrations combine the distributed memory and computation with analog signals that give brains some significant survivability advantages.

Recently developed Contrastive Local Learning Networks (CLLNs) \citep{dillavou_machine_2024} may fill this gap. A network of self-adjusting nonlinear resistors, these systems perform supervised learning in an emergent manner by utilizing physical processes to perform computation. Learning occurs in each individual element using local rules, eschewing central and digital processing. As a result, they are energy efficient \citep{stern_training_2024} and fault tolerant \citep{dillavou_demonstration_2022}. However, CLLNs have not been demonstrated in an RL context, even in simulation. 

\begin{figure}[ht]
    \begin{center}
     \includegraphics[width=\linewidth]{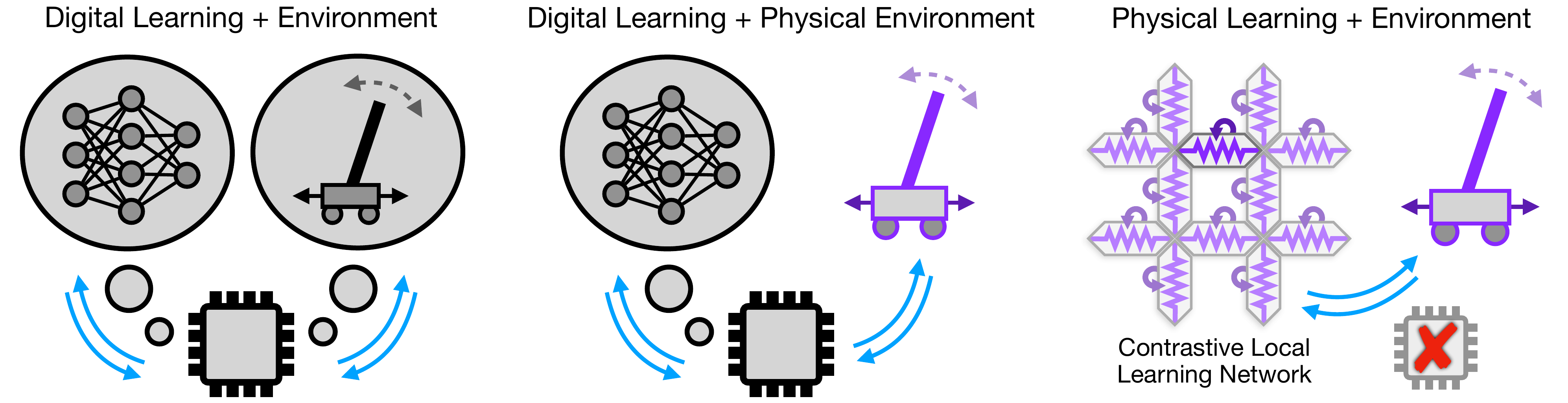}
    \end{center}
    \caption{A schematic of reinforcement learning in three scenarios. Left: A digital agent learns in a digital environment, both simulated by a computer. Middle: A digital agent interacts with a physical environment, and the learning process is controlled by a computer. Right (proposed): A Contrastive Local Learning Network interacts with a physical environment, and learns based on those interactions. There is no digital processing; learning is done in a distributed, analog fashion. One of the many identical self-adjusting components is highlighted.}
    \label{fig:fig1}
\end{figure}

Designing an RL implementation of CLLNs provides a setting to ask a range of questions. For instance, which RL tools are natural for a self-learning network, and which require additional pre-programmed hardware? There are numerous challenges associated with digital RL acting in the real (non-simulated) world \citep{dulac-arnold_challenges_2021}; what additional challenges stem from placing the agent's brain outside of the pristine digital realm (Fig.~\ref{fig:fig1})? A successful navigation of these challenges would open a path towards energy-efficient and fault-tolerant autonomous agents.

\section{Reinforcement Learning with a Simulated CLLN}

\subsection{Background: Contrastive Local Learning Networks}
Contrastive Local Learning Networks (CLLNs) are networks of self-adjusting resistive elements whose individual dynamics approximate gradient descent on a global loss function \citep{dillavou_machine_2024}. The standard operation of these systems is for supervised learning and is briefly summarized here.

\begin{figure}[ht]
    \begin{center}
     \includegraphics[width=0.7\linewidth]{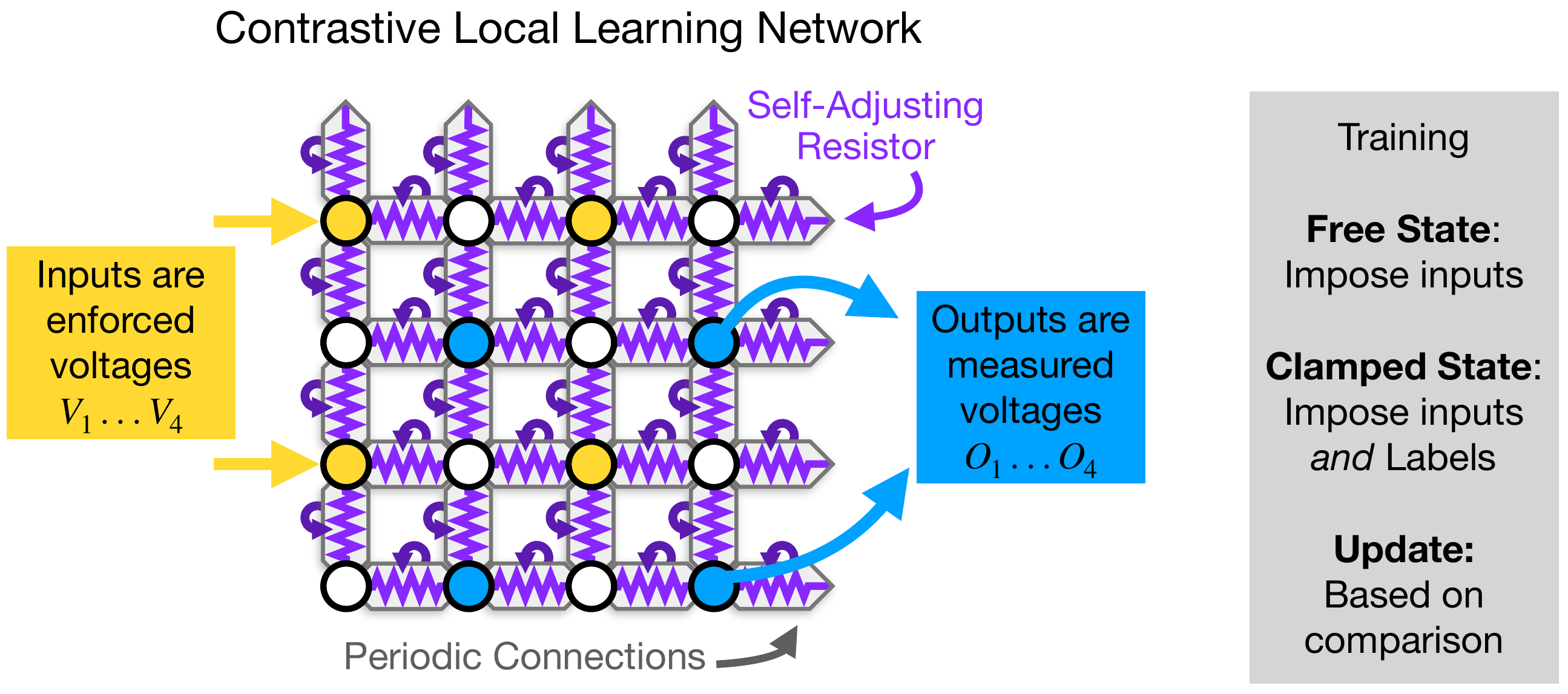}
    \end{center}
    \caption{Schematic of a Contrastive Local Learning Network (CLLN). The configuration shown is used for the Markov decision process with four states and four actions. A modified network architecture and input-output position is used for the 9-state navigation task (see Fig.~\ref{fig:fig4}). A high-level description of the contrastive training protocol is outlined in the gray box. Note that each element in a CLLN requires only local measurements of the two states (free and clamped) to update itself, decentralizing the training process.}
    \label{fig:fig2}
\end{figure}

\renewcommand{\thefootnote}{2}
We consider a network of resistors as shown schematically in Fig.~\ref{fig:fig2}. We will treat four of the nodes (yellow) as inputs, and encode data by enforcing the voltage values of these nodes ($V_1, \dots,V_4$). We will also choose four nodes (blue), and treat their equilibrium voltage values, $O_1, \dots, O_4$, as outputs. That is, we treat our network as a physical function, $\mathcal{F}(V_1, V_2, V_3, V_4) \equiv (O_1, O_2, O_3, O_4)$. It is a consequence of our eventual choice of task that the number of inputs and outputs is equivalent; they are independently specified in general. When input voltages are applied, output values are `calculated' by the physical processes, in this case the flow of current in the network. The result of this calculation is a consequence of the network structure and the conductance (inverse of resistance) values of each edge of the network, $G_i$. We note that at equilibrium, the voltages at unconstrained nodes have minimized power dissipation with respect to the boundary conditions\footnote{In a nonlinear network, like the one we consider, the true minimized quantity is \textit{co-content} which reduces to power in the linear case. As this is a less-commonly known quantity we use power throughout. This does not change the intuition or derivations.}.

The evolution of the gate voltages (the trainable parameters) in this network is \textit{contrastive} in that it requires comparing two distinct states that we can impose. Given a single data point (inputs $V_1 \dots V_4$) and label,  (desired outputs $L_1 \dots L_4$), the two states are as follows. First, the \textit{free} state, wherein only inputs $V_1, \dots, V_4$ are imposed. In this state, each resistor experiences voltage drop $\Delta V^F_i$ and the outputs find values $O^F_n$. We use subscript $i$ to denote quantities on network edges, and $n$ to denote quantities on nodes (inputs or outputs). Next, the \textit{clamped} state, in which inputs as well as the desired outputs (labels) are enforced. Now each resistor experiences voltage drop $\Delta V^C_i$, and the (enforced) outputs are equal to the labels $O^C_1 = L_1$, etc. In practice the clamped output is actually nudged \textit{towards} the label: $O^C_n = O^F_n (1-\eta) + \eta L_n$, where $\eta$ is a hyperparameter, and $\eta=1$ simply clamps the label ($O^C_n = L_n$). In this work we use $\eta = 0.1$.

The system will perform gradient descent on a \textit{contrastive} function, with respect to its conductance values, following the Coupled Learning framework \citep{stern_supervised_2021}. This framework is closely related to Equilibrium Propagation \citep{scellier_equilibrium_2017} and Contrastive Hebbian Learning \citep{movellan_contrastive_1991}. Specifically,
\begin{equation} \label{graddesc}
    \delta G_i = -\alpha \frac{d}{dG_i} \Big[\underset{\text{contrastive fn}}{\mathcal{P}^C-\mathcal{P}^F}\Big],
\end{equation}
where $\mathcal{P}^C$ and $\mathcal{P}^F$ are the total dissipated power in the clamped and free states respectively, and $\alpha$ is a learning rate. We note that because the electronic network minimizes power with respect to its boundary conditions, and the clamped state is equivalent to the free state with added boundary conditions (clamped outputs), the contrastive function must be non-negative. Further, it is only zero-valued when the clamping procedure does not change the outputs, that is, the labels are already satisfied. Thus, it has the same minima as a mean-squared error cost function, and in practice its minimization works well to minimize error \citep{stern_supervised_2021, dillavou_machine_2024, dillavou_demonstration_2022}.

Calculating the derivative on the RHS of eq.~\ref{graddesc} for the clamped power with the chain rule, we find that
\begin{equation} \label{makelocal}
   \frac{d\mathcal{P}^C}{dG_i} = \sum_j^{\text{edges}}\frac{\partial \mathcal{P}^C}{\partial G_j}\cancelto{\delta_{ij}}{\frac{dG_j}{dG_i}} + \cancelto{0}{\frac{\partial \mathcal{P}^C}{\partial \Delta V^C_j}}\frac{d\Delta V^C_j}{dG_i} =\frac{\partial \mathcal{P}^C}{\partial G_i}.
\end{equation}
The first term cancels for $i\neq j$ as the conductances are independently set, and the second cancels as power is (as noted) at a minimum with respect to the voltages. As a result, the total derivative becomes a partial derivative, which we can further simplify by noting that power is a sum over local powers:
\begin{equation} \label{localderiv}
    \mathcal{P}^C = \sum_j^{\text{edges}} \left(\Delta V_j^C\right)^2 G_j 
    \quad \rightarrow \quad 
    \frac{d\mathcal{P}^C}{dG_i}  = \frac{\partial \mathcal{P}^C}{\partial G_i} = \left(\Delta V_i^C\right)^2 
\end{equation}
Thus, we complete the same calculation for the free power and the result is a local learning rule,
\begin{equation} \label{localrule}
    \delta G_i = -\alpha \frac{d}{dG_i} \Big [\underset{\text{contrastive fn}}{\mathcal{P}^C-\mathcal{P}^F}\Big ] = \alpha \Big [\left(\Delta V_i^F\right)^2-\left(\Delta V_i^C\right)^2\Big]
\end{equation}
Wherein each element needs only to measure its own voltage drop in both states to enact this gradient descent. In practice, this is done by creating twin networks co-evolving networks, one holding the free state, the other the clamped state \citep{dillavou_demonstration_2022,dillavou_machine_2024}. 

We model our simulations after \citet{dillavou_machine_2024}, where each conductive element is in fact a MOSFET transistor in the triode (passive) regime, with conductance equation
\begin{equation} \label{conductance}
    G_i = S(V_{G,i}-V_T- \overline V )
\end{equation}
where in our simulations $S=1$ and $V_T=0.7$ are constants, $V_{G,i}$ is the (adjustable) gate voltage that will act as weights in our system, and $\overline V$ is the average of the node voltages on either side of the edge. This final term allows the system to perform nonlinear input to output transformations. We note that because $\frac{\partial G_i}{\partial V_{G,i}} = S$, the gradient with respect to $G_i$ and $V_{G,i}$ are co-linear, and we may enact eq~\ref{localrule} by changing $V_{G,i}$, which will be our learning procedure. We restrict the range of our parameters such that $1.0 < V_{G,i} < 5.5$ to mimic experimental conditions.

\subsection{Example 1: A toy Markov decision process with four states and four actions}
We now adapt this scheme for Q-Learning on a Markov decision process (MDP) with four states and four actions. We will treat our CLLN simulation as enacting the Q matrix. At each training step $t$, we encode the environmental state $S_t$ as the input; states $S_1 \dots S_4$ are encoded as [$V_1,V_2,V_3,V_4$] = [1 0 1 0], [0 1 0 1], [1 1 0 0], and [0 0 1 1] V respectively. Note that all voltages are analog, despite the input appearing to be digital, and the outputs will fall between 0 and 1. We then use epsilon-greedy action selection where $\epsilon$ decays from 0.05 to 0 linearly over the training run. We select the maximum output of the four (with probability $1-\epsilon$) as our action $A_t$.

\begin{figure}[b]
    \begin{center}
     \includegraphics[width=0.7\linewidth]{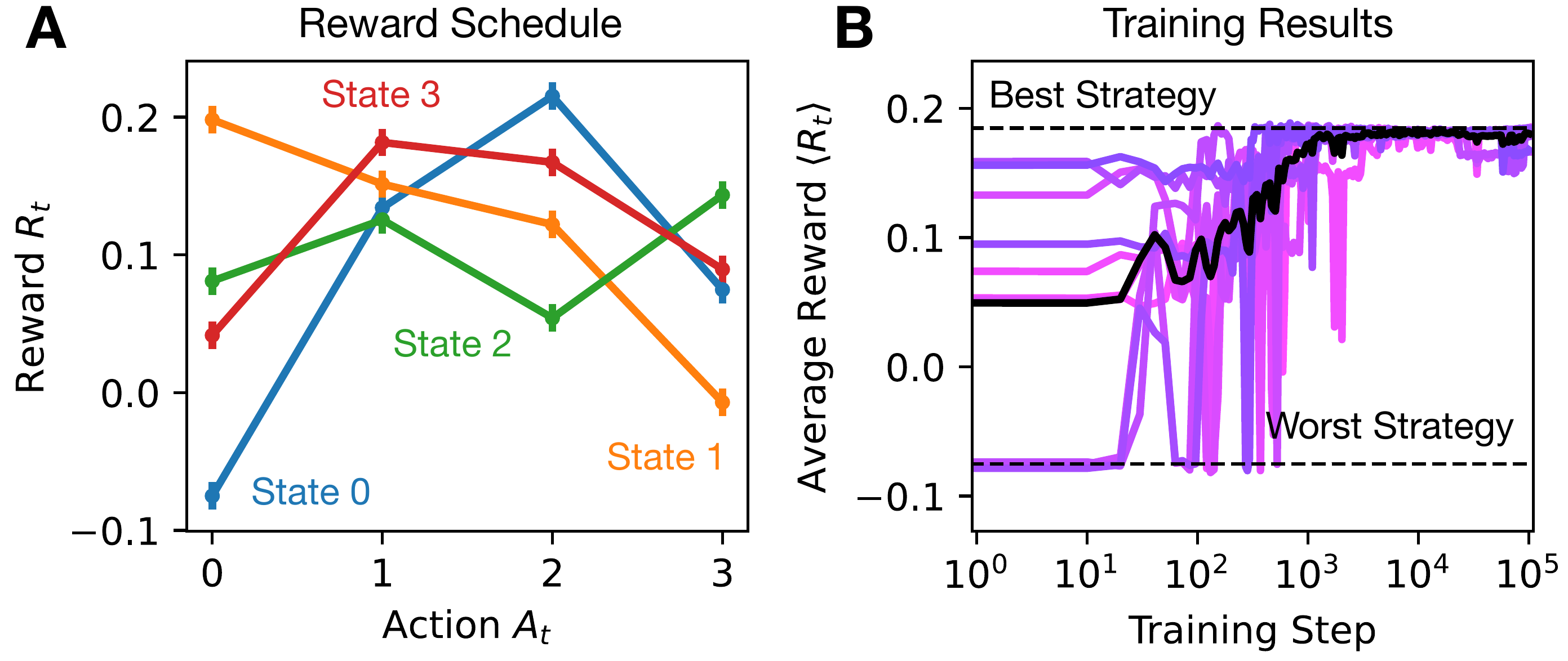}
    \end{center}
    \caption{Q Learning with CLLN. (A) Reward schedule for each state (noise effect shown as small error bars). The optimal strategy involves a cycle through all four states. (B) Average reward over training for 10 trials (purples) overlaid with their average (black).}
    \label{fig:fig3}
\end{figure}

The environment then issues a state-dependent reward $R(S_t,A_t) + \mathcal{N}(0,0.01)$ and the state changes to $S_{t+1}$. $R(S_t,A_t)$ was sampled from $\mathcal{N}(0.1,0.1)$, and the selected values are shown in Fig.~\ref{fig:fig3}A. Our environment is extremely simple, with action 0 inducing state 0, action 1 inducing state 1 and so on. We then impose the inputs for state $S_{t+1}$, and take the maximum output minus the mean of all four outputs as our predicted future reward. We use a discount factor $\gamma=0.5$ and all of this information to create future-weighted score $L_t$:
\begin{equation}
    L_t = R(S_t,A_t) + \gamma \left [ \max(\mathcal{F}(S_{t+1}))- \text{mean}(\mathcal{F}(S_{t+1})) \right ]
\end{equation}
We the subtracted mean term improves performance, as it removes a constraint on the average set out outputs, permitting our small network additional flexibility. We allow the system to evolve (eq~\ref{localrule}) while imposing encoded $S_t$ as inputs, $O_i$ on all clamped-state outputs corresponding to actions the system \textit{did not} take, and $L_t$ on the clamped-state output whose action was taken. 

As an example: the system is in state $S_t=1$, and $O_2$ is the biggest (we assume we do not take the rare random action). The agent takes action $A_t = 2$ as a result, gets reward $R_t$, and the state moves to $S_{t+1} = A_t = 2$. $L_t$ is calculated as above, and then we allow our system to evolve (eq~\ref{localrule}) while $S_t$ is imposed as input in both free and clamped states, and $\mathcal{F}(S_{t+1})$ is imposed as the clamped label with $L_t$ inserted at position 2 (the action taken). That is, we train the system to keep the outputs of the untouched actions the same, while moving the chosen action's output value towards $L_t$. The result is evolution similar to the Bellman equation. Updates are batched and imposed on the system every 50 steps.

\subsection{Results}

We perform 10 trials of 100,000 training steps, each time randomly initializing $V_{G,i}$ using $\mathcal{N}(1.5,0.1)$. Rewards are tallied over time, and averaged in logarithmically spaced intervals (minimum of 10 steps) for each trial, shown as purple lines in Fig.~\ref{fig:fig3}B, with their average overlaid black. We simulate rewards for our same protocol (with $\epsilon=0$ but including reward noise) using 10,000 steps with each of the $4^4=256$ possible strategies, and denote the best and worst cases with black dashed lines in Fig.~\ref{fig:fig3}B. Our system always learns a near-optimal strategy, including ending at the best possible strategy in 8 of 10 trials.

\subsection{Example 2: Navigation on a grid with nine states and four actions}
\begin{figure}[b]
    \begin{center}
     \includegraphics[width=0.9\linewidth]{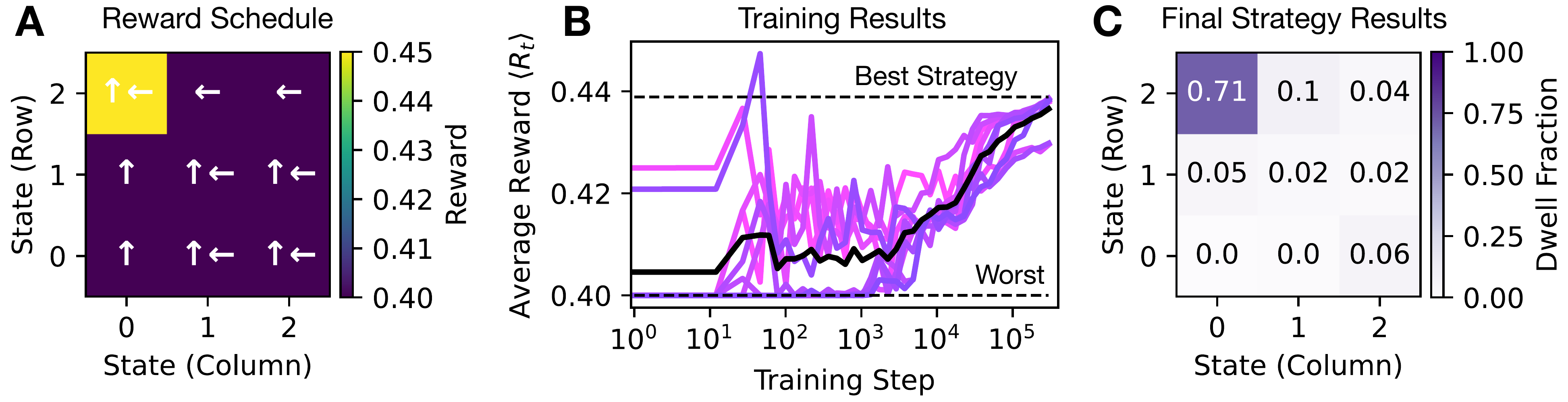}
    \end{center}
    \caption{Navigation Task. (A) Reward schedule, with all states approximately equal except for the target (upper left) state. There is no reward noise. The optimal action in each grid is indicated via arrows. (B) Average reward over training for 10 trials (purples) is overlaid with their average (black). (C) The 10 strategies at the end of training are each simulated for 10,000 steps (reset randomly every 5 steps), and the fraction of total time spent in each state is shown as a heatmap. }
    \label{fig:fig4}
\end{figure}

We use a modified network architecture to attempt a slightly more difficult task, navigating towards a goal location in discretized 2D space. The reward schedule is shown in Fig~\ref{fig:fig4}A, with a reward gradient (shaping) 5000 times smaller than the large reward (top left), invisible on the heatmap. The optimal strategy (overlaid) is to always move up or left.

We now simulate a larger, 44-edge network to perform this task. The architecture is equivalent to a densely connected neural network, with 6, 4, 4, and 1 node in each layer. Note however that our network is \textit{not} feed forward, and thus we place our four output nodes in the third layer, and have each one represent a move (up, down, left, right). The row and state values are rescaled [0, 0.5, 1] and imposed as two inputs in the first layer. Their inverses [1, 0.5, 0] are also imposed in the first layer, and the final two nodes in this layer are always set to 0 and 1, respectively. The single node in the final layer is set to 0.5, making 7 total inputs, only 4 of which change (with the state). When training (and testing) the system is randomly placed in a new state every 5 steps.

\subsection{Results}
As before, we randomly initialize our network ($V_{G,i}$ from $\mathcal{N}(1.5,0.1)$) and batch updates every 50 training steps. We train 10 trials for 300,000 steps, reduce $\epsilon$ from 0.1 to 0 linearly in each trial, and report the average rewards for each section of training in purple in Fig.~\ref{fig:fig4}B, with their average overlaid in black. In 8 of our 10 trials, our system finds one of the optimal strategies (several are equivalent). Note that the system cannot sit on the 'best strategy' line while $\epsilon > 0$, which is the case until the very end of training.

After training, we simulate each trained agent operating in the environment for 10,000 steps, still randomly setting their location every 5 steps, but with no random actions beyond this ($\epsilon = 0$). We find the agents spend most of their time in the high-reward square, as shown in Fig~\ref{fig:fig4}C.

\section{Discussion: Implications of reinforcement learning on a physical network}

The reinforcement learning (RL) scheme developed for a physical network performs well on an MDP comprising four states and four actions, and on a navigational task in a two dimensional grid, with nine states and four actions. Trained agents find optimal strategies with few exceptions. While the above algorithm is quite simple, its development required engaging with two atypical features when compared to digital RL:
\begin{itemize}
    \item \textbf{Parameters and outputs in our system are bounded.} As a result, we must take care not to define rewards or discount factors such that the system cannot produce the required outputs.
    \item \textbf{CLLNs are not `feed forward'.} A voltage anywhere in the network (or a change in resistance anywhere) potentially impacts every output. Thus, our protocol required us to train unselected outputs to remain stationary. The limitations this poses on the complexity of agent strategies is a subject for future study, as are the advantages of a network capable of sensing inputs from any location.
\end{itemize}

Further, envisioning a physical system performing our scheme necessitates a somewhat `unnatural' step backwards in time:
\begin{enumerate}
    \item System evaluates state $S_t$ and selects action $A_t$ 
    \item Environment returns reward $R_t$, and switches to $S_{t+1}$
    \item System evaluates state $S_{t+1}$ and calculates $L_t$
    \item System updates based on \textbf{previous} state $S_t$ and $L_t$
\end{enumerate}
This requires storing and reapplying $S_t$ after the environment has moved on ($S_{t+1}$). It is possible that some constructed hybrid state (involving $S_{t+1}$ and $L$) could allow the system to avoid such `backtracking' and still train successfully. 

While not necessary for our simple tasks, it is not obvious how to naturally introduce history-storage methods like eligibility traces or replay buffers  ~e.g.~\citep{haarnoja2018soft, schulman2017proximal, mnih2013playing}. Utilizing these tools with how CLLNs evolve would require extensive control hardware, moving away from the vision of a `network in the wild'. However, even our current simple implementation required some external control, to implement the randomization and the max function in the epsilon-greedy algorithm, and to hold and swap back $S_t$ as discussed above. While a future CLLN could certainly be trained to perform these functions, the system then is being asked to learn the learning algorithm before succeeding at the task. Other schemes in the RL literature are far more natural to include. For instance, implementing both a policy function and value function could be done with a single network, or by dividing the network in two. 

By virtue of being a simulation, our CLLN has several features of the digital world. However when physically constructed, these aspects change. First, in hardware, imperfections in components and measurements can manifest as biased elements, hamstringing the learning process and preventing capture of subtle learning signals~\citep{dillavou_understanding_2025}. Such effects are avoided (for good reason) in RL and machine learning writ large, but spotlight a distinct advantage that these algorithms have over physically instantiated learning systems like the brain.

Second, the notion of `damage' is somewhat meaningless in simulation, but has very real consequences for biological systems that rely on a physical substrate. As CLLNs are collections of self-adjusting elements and can retrain after damage \citep{dillavou_demonstration_2022}, developing architectures and algorithms to allow on-the-fly recovery in RL tasks is an exciting research direction, and one that makes little sense in the digital domain. 

Third, the energy used by the processor in simulating our CLLN is independent of the resistances and currents \textit{in silico}. However a physically constructed version burns power directly proportional to these values, and small modifications to the learning have been demonstrated to bias physically instantiated CLLNs towards low power solutions \citep{stern_training_2024}. Perhaps networks could likewise be biased into states more robust to physical damage, or with faster input-output transmission. Modifications to learning algorithms that accounts for such secondary goals are an exciting avenue for future work, and as above, make little sense in digital systems.

\section{Conclusion}

This work highlights a new type of analog, distributed system, a Contrastive Local Learning Network (CLLN), whose features make it a natural fit for reinforcement learning (RL). We simulated two CLLNs, and developed a training scheme to leverage their self-learning dynamics to enact Q Learning on one Markov decision processes each, the first with four states and four actions, the second with a nine states ($3 \times 3$ grid) and four actions (cardinal direction moves). The CLLN agents were largely successful, achieving near optimal rewards across trials. Our scheme highlighted several aspects of typical RL algorithms that are less natural to apply to physical self-learning networks, most including some form of additional nontrivial memory. Finally, we discussed disadvantages like imperfect components and secondary goals such as low-power operation and robustness to damage, all of which may be improvable via modified training methods in CLLNs, a process that makes little sense in the digital domain.

\subsubsection*{Acknowledgments}
\label{sec:ack}
This work is supported by funding from the UK Advanced Research + Invention Agency (ARIA) under project NACB-SE02-P06

\bibliography{bib}
\bibliographystyle{rlj}

\appendix

\end{document}